\documentclass{article}

\usepackage[final]{format}

\usepackage[utf8]{inputenc} 
\usepackage[T1]{fontenc}    
\usepackage{hyperref}       
\usepackage{url}            
\usepackage{booktabs}       
\usepackage{amsfonts}       
\usepackage{nicefrac}       
\usepackage{microtype}      
\usepackage{caption}
\usepackage{multirow}
\usepackage{graphicx}
\usepackage{caption}
\usepackage{subcaption}
\usepackage{amsmath}

\graphicspath{ {./images/} }
\title{Geometric-informed GFlowNets for Structure-Based Drug Design}

\author{
  Grayson Lee \\
  Simon Fraser University\\
  \texttt{graysonl@sfu.ca} \\
  \And
  Tony Shen \\
  Simon Fraser University\\
  \texttt{tsa87@sfu.ca} \\
  \And
  Martin Ester \\
  Simon Fraser University\\
  \texttt{ester@sfu.ca} \\
}
\begin{document}

\maketitle

\begin{abstract} 
The rise of cost involved with drug discovery and current speed of which they are discover, underscore the need for more efficient structure-based drug design (SBDD) methods. We employ Generative Flow Networks (GFlowNets), to effectively explore the vast combinatorial space of drug-like molecules, which traditional virtual screening methods fail to cover. We introduce a novel modification to the GFlowNet framework by incorporating trigonometrically consistent embeddings, previously utilized in tasks involving protein conformation and protein-ligand interactions, to enhance the model's ability to generate molecules tailored to specific protein pockets. We have modified the existing protein conditioning used by GFlowNets, blending geometric information from both protein and ligand embeddings to achieve more geometrically consistent embeddings. Experiments conducted using CrossDocked2020 demonstrated an improvement in the binding affinity between generated molecules and protein pockets for both single and multi-objective tasks, compared to previous work. Additionally, we propose future work aimed at further increasing the geometric information captured in protein-ligand interactions.

\end{abstract}

\section{Introduction} 

Drug design is an increasingly expensive procedure, a trend underscored by Eroom's law, which highlights how the cost of developing new drugs has exponentially risen over recent decades, consequently slowing down development \cite{Scannell2012}. At the heart of addressing these challenges is Structure-Based Drug Design (SBDD), a method that involves considering the structure of the protein receptor during drug development. Typically, SBDD employs virtual screening, a process that predicts the interactions between a protein pocket and a virtual library of molecules through protein-ligand docking. However, this method faces limitations due to the exhaustive nature of evaluating every possible interaction and the size constraints of the virtual library.

To enhance the efficiency of SBDD, machine learning techniques have been applied to predict protein-ligand docking outcomes. These methods aim to streamline the screening process by intelligently navigating the combinatorial space of drug-like molecules, which is estimated to be on the order of $10^{60}$ \cite{Bohacek1996StructureBased}, a stark contrast to the available in-silico data on protein-ligand interactions, such as the PDDBind database, which includes around $10^4$ structures \cite{Wang2005PDBbind}.

Furthermore, an emerging approach within SBDD is the use of generative models that leverage protein-ligand datasets to generate molecules fitting specific protein pockets. Traditional models have employed autoregressive 3D-graph neural networks, while methods, like those proposed by Guan et al.\cite{Guan20233DEquivariant}, utilize diffusion processes to generate molecules in 3D. These generative approaches have demonstrated the ability to achieve high docking scores and favorable drug likelihood metrics, such as Quantitative Estimate of Druglikeness (QED) and Synthetic Accessibility (SA). However, despite fitting the distribution of known interactions, these models often struggle to outperform the datasets they are trained on and face challenges in generalizing to unseen protein pockets.

In addressing the limitations inherent in traditional structure-based drug design (SBDD) methodologies—particularly their inadequate exploration of chemical space—reinforcement learning methods have been leveraged. Notably, Generative Flow Networks (GFlowNets) \cite{bengio2023gflownet} are distinguished by their theoretical guarantees, which ensure the generation of diverse sets of candidates in proportion to their associated rewards. This characteristic is crucial in ligand generation for specific protein pockets, where the diversity of the compounds directly impacts the success of SBDD efforts.

Significant progress in this domain has been highlighted by the contributions of Shen et al. \cite{Shen2023TacoGFN}, who have explored the incorporation of protein pocket conditioning within the GFlowNet architecture. Through adjustments to the reward function, their approach has demonstrated enhanced performance in achieving desirable drug qualities, such as Quantitative Estimate of Druglikeness (QED), Synthetic Accessibility (SA), and predicted docking scores, outperforming traditional models reliant on distribution fitting for these metrics.

Despite these advances, significant information is lost due to the choice of encoding within a 2D space, as the process of encoding the 3D protein pocket into a conditioning embedding is insufficient. The diversity observed across various protein pockets has been seen to be lower than that in distribution models. This underscores a crucial area for further research, aiming to refine the use of GFlowNets in the drug discovery process. A pivotal challenge lies in augmenting the ability of GFlowNets to more effectively encode spatial pocket information, a step that could substantially improve their ability to generate pocket-specific molecules.

In this paper, we introduce a modification to the existing framework for GFlowNet pocket embedding, drawing from the work of Zhang et al. \cite{zhang2023e3bind}, who demonstrated the efficacy of incorporating spatial geometric constraint between proteins and ligands for in-silico docking. To enhance it's ability to contextualize protein pockets, we adapt the transformer architecture to include the protein and ligand node distances, in addition to their possible interactions. This modification aims to preserve geometry awareness between fragment-residue pairs, thereby enriching the pocket conditioning process with more detailed spatial information.

Our contributions include the following:
\begin{itemize}
\item We enhance pocket conditioning by incorporating geometric information from protein-ligand interactions, using intra-protein and intra-pocket distances.
\item We surpass previous work by using a geometry-enriched protein-ligand embedding to improve the binding affinity between the target protein pocket and generated molecules.

\end{itemize}

\section{Related Works} 


\paragraph{Structure-based drug design} conditions on protein structure information in order to better generate molecules. This provides a more generalize model which can adapt to unseen pockets. Common approaches to leveraging this conditioned information include Pocket2Mol\cite{Peng2022Pocket2Mol} and 3DSBDD\cite{Luo20223DGenerativeModel} which sample pockets in 3D and use an autoregressive method to iteratively build the molecule from atoms. Another approach taken by FLAG\cite{Zhang2023MoleculeGeneration} and DrugGPS\cite{Zhang2023LearningSubpocket} is to instead build molecules from common molecular fragments to leverage existing chemical information and reduce the search space. Diffusion based models such as TargetDiff\cite{Guan20233DEquivariant} and DecompDiff\cite{pmlr-v202-guan23a} use diffusion to generate molecules in the 3D space.

\paragraph{GFlowNets} use a trained stochastic policy in order to generate an object through a sequence of steps. The key is that it is able to generate a diverse set of objects with a probability proportional to the reward function. GFlowNets have been used by Shen et al\cite{Shen2023TacoGFN} to be able to generate molecules conditioned on pocket information, with desirable drug-like qualities such as SA and QED, through the use of multi objectives\cite{Jain2023MultiObjectiveGFlowNets}. However, they suffer from low inter-pocket diversity compared to distribution-based models. GFlowNets promise a more diverse set of generated molecules. This may be due to their generation of ligands in 2D space, which enables them to generate molecules more rapidly than 3D methods. However, this approach has the downside of losing information from the 3D protein pocket.

\paragraph{Protein-ligand interaction} has played a crucial role in advancing the modeling of SBDD. A key development in this area is the use of triangle attention mechanisms, which maintain the geometric consistency of the distance map, improving the representation of molecular structures. Jumper et al. \cite{Jumper2021} were among the first to apply distance geometry to the challenge of protein folding, conceptualizing it as a graph inference problem within three-dimensional space. Their approach, which generates pairs of proximate atoms, enables the prediction of protein structures by creating a 3D conformation maps. Expanding on these concepts, Lu et al. \cite{Lu2022TANKBind} proposed a triangle attention mechanism tailored for protein and ligand atom interactions. Diverging from conventional conformation sampling approaches, their method predicts the protein-ligand distance map directly. This map is then converted into a docked pose using gradient descent, with an optimization function that minimizes the discrepancies in protein-ligand and intra-ligand distances from their predicted and actual values. Zhang et al. \cite{zhang2023e3bind} further improved upon these methodologies by calculating geometry-informed attention weights for neighboring atom pairs, based on intra-protein distance embeddings for each attention head. Their technique provides a more detailed evaluation of the interactions between atoms, contributing to the refinement of molecular dynamics and interaction.

\section{Method} 
\paragraph{}
We address the problem of protein pocket encoding by introducing geometric information into the final embeddings, through a geometry-aware Trioformer \cite{zhang2023e3bind}. Which combines intra-ligand and intra-protein node distances to learn attention weights for between embeddings. 
\subsection{Preliminaries}
\subsubsection{Protein Encoding}
Shen et al. \cite{Shen2023TacoGFN} employ the GFlowNet architecture to encode the protein pocket, denoted as $p$, using a $K$-nearest neighbor (KNN) graph with standard euclidean distance. The graph $\mathcal{G}^\mathcal{P}$ contains nodes $v_i^{\mathcal{P}} \in \mathcal{V}^{\mathcal{P}}$ where they contain molecular information such as the type of residue, dihedral angles, and directional unit vectors. The edges $e_{ij}^{\mathcal{P}} \in \mathcal{E}^{\mathcal{P}}$ are any edges such that $v_j$-th node is in $v_i$'s $K$-nearest neighbors. As the edges intend to encode distance they contain information of euclidean distance, distance along the backbone, and a direction vector between the nodes. This KNN graph contains sufficient spatial information as the graph should be equivariant to rotation and 3D translation. Geometric vector perceptrons introduced by Jing et al.\cite{jing2021learning} are then used, due to their equivariance of rotation and 3D translation, to encode $\mathcal{G}^\mathcal{P}$ into a set of node embeddings $\{h_{{v_i}^p}\}$ . We then averaged them to get a single embedding $h_{\mathcal{G}^\mathcal{P}}$ for the protein pocket graph, of which the GFlowNet will use for conditioning.

\subsubsection{Ligand Encoding}
\paragraph{}
Shen et al. \cite{Shen2023TacoGFN} encode the ligand, denoted as $L$ as a 2D molecular graph. The graph $\mathcal{G}^L = (\mathcal{V}^L, \mathcal{E}^L)$, consists of nodes $v^L_i \in \mathcal{V}^L$ which contain a molecule fragment. The edges $e^L_{ij} \in \mathcal{E}^L$ represent the attachment atoms within the fragments $v^L_i$ and $v^L_j$, where possible attachment points are specified in the fragment library.

\subsubsection{Pocket Conditioning}
The graph transformer \cite{yun2020graph} is employed to model the GFlowNet policy $\pi(\mathbf{a}|\mathcal{G}^L_t, \mathbf{h}_{\mathcal{G}^\mathcal{P}})$ at each time step $t$, using the current 2D ligand molecular graph alongside the pocket graph embedding $\mathbf{h}{\mathcal{G}^\mathcal{P}}$. Initially, the ligand graph nodes $\mathbf{h}i^{L(0)}$ are represented through a one-hot encoding of their fragment type $v^L_i \in \mathcal{V}^L$. Similarly, the edges $\mathbf{e}{ij}^{L(0)}$ are denoted by a one-hot encoding of the attachment atom index from $v_i^L$ to $v_j^L$. To incorporate the pocket graph embedding, they extend the ligand graph by introducing a virtual node $v_k^L$ and augment $\mathcal{E}^L$ to include edges $e_{ki}^L$ for each node $v_i^L$ within $\mathcal{V}^L$, thereby connecting the new virtual conditioning node to every existing node in the ligand graph. This augmented ligand graph, is then passed through $N$ layers of the graph transformers. After processing, the virtual node $\mathbf{h}_k^{L(N)}$ is separated from the augmented graph.

The final node embeddings ${\mathbf{h}_i^{L(N)}}$, excluding the virtual node, are then subjected to global average pooling and then concatenated with the final virtual node embedding $\mathbf{h}^{V(N)}$ to obtain a graph-level embedding $\mathbf{g}^{L(N)}$. Mathematically, this is represented as:
\ref{eq: graph embedding}:
\begin{equation}
    \mathbf{g}^{L(N)} = Concat\left(AvgPool\left(\left\{\mathbf{h}^{L(N)}_i\right\}\right), \mathbf{h}^{V(N)}\right)
    \label{eq: graph embedding}
\end{equation}
Subsequently, edge embeddings $\mathbf{e}_{ij}^{L(N)}$ for each edge in $\mathcal{E}^L$ are calculated by adding the final node embeddings of the connected nodes.
\ref{eq: edge embedding}:
\begin{equation}
    \mathbf{e}_{ij}^{L(N)} = \mathbf{h}^{L(N)}_i + \mathbf{h}^{L(N)}_j
    \label{eq: edge embedding}
\end{equation}
Thus, the gflownet utilizes three types of embeddings for its operation: node embeddings ${\mathbf{h}_i^{L(N)}}$, edge embeddings ${\mathbf{e}_{ij}^{L(N)}}$, and the graph-level embedding $\mathbf{g}^{L(N)}$, which collectively contribute to modeling the policy $\pi$. We can see that while the gflownet conditions on the protein pocket embedding, that there is a significant lost of information as the protein information is encoded into a single vector.

\subsection{Encoding geometric representations from the pocket}
In our approach, we take follow Zhang et al. \cite{zhang2023e3bind} by encoding additional spatial information between the protein-ligand interactions. We use the original KNN graph of the protein structure and instead of passing in a protein pocket graph embedding vector to the graph transformer we pass in a protein pocket node embeddings in addition to a distance matrix of each atom within the pocket. We also compute the a neighbours matrix for the ligand graph. By using these two matrices we are able to provide additional geometric information between the pocket and ligand. Unlike previous work with distance geometry \cite{zhang2023e3bind} we do not have access to 3D coordinates of the ligand as we have a 2d graph, however we can take into account the most important information which is each nodes neighbours within the ligand graph. We thus create an adjacency matrix of the neighbours.

\subsection{Geometric Informed Embedding}
In our approach, we initially encode the protein pockets and ligands in alignment with existing methodologies, utilizing a geometric vector perceptron for proteins and a graph transformer for ligands. Unlike prior models that concatenate pocket encoding directly onto the ligand embeddings, we enhance the representation of geometric relationships by leveraging the embedding vectors alongside 3D coordinates of the pocket. Specifically, we utilize an adjacency matrix to represent the proximity between ligand fragment neighbors, differently from Zhang et al. \cite{zhang2023e3bind}, who utilized ETKDG conformations \cite{riniker2015better}. In the absence of explicit conformations, our method calculates proximity using the adjacency matrix, to provide a simple approximate of adjacent connection distances through one-hot encoding.

We adopts a Trioformer architecture from Zhang et al. \cite{zhang2023e3bind} to iteratively update and combine the protein, ligand, and pair embeddings. Our trioformer block takes the initial protein node embeddings \( \{h_{{v_i}^p}\} \) and ligand node embeddings \( h_i^{L(N)} \), and constructs a set of pairwise embeddings \( h_{ij}^{PL} \) that capture the interactions between each protein and ligand node pair. To effectively integrate the spatial context of both intra\-protein and intra\-ligand the pair embedding are informed by the distances between nodes, specifically ligand distances \( d_{ik}^{L} \) and protein distances \( d_{jk'}^{P} \), encapsulated within the embeddings as attention biases.

For each protein-ligand pair, we calculate an attention bias using a linear projection of the distance embeddings \( t_{ij}^{(h)} = \text{Linear}^{(h)}(d_{ij}^{\ast}) \), where \( \ast \) represents either intra-ligand or -intra-protein distances and \( h \) indicates the attention head. This projection serves to update the attention scores, thereby encoding the spatial constraints directly into the attention process.

Incorporating these biases, the attention scores for each head are computed as $( a_{ijk'}^{(h)} = \text{softmax}(\frac{1}{\sqrt{c}}(q_{ij}^{{(h)}^T }k_{ik'}^{(h)} + b_{ij}^{(h)} + t_{jk'}^{(h)})) )$, where $( q_{ij} )$, $( k_{ik'} )$ and $b_{ij}^{(h)}$ are the query, key and bias matrices derived from the node pair embeddings, and $( t_{jk'}^{(h)} )$ is a distance-based bias specific to each head. These scores then weigh the aggregation of neighboring information, ensuring that closer node pairs have a greater influence on the updated embeddings.

\begin{figure}
     \centering
         \includegraphics[width=8cm]{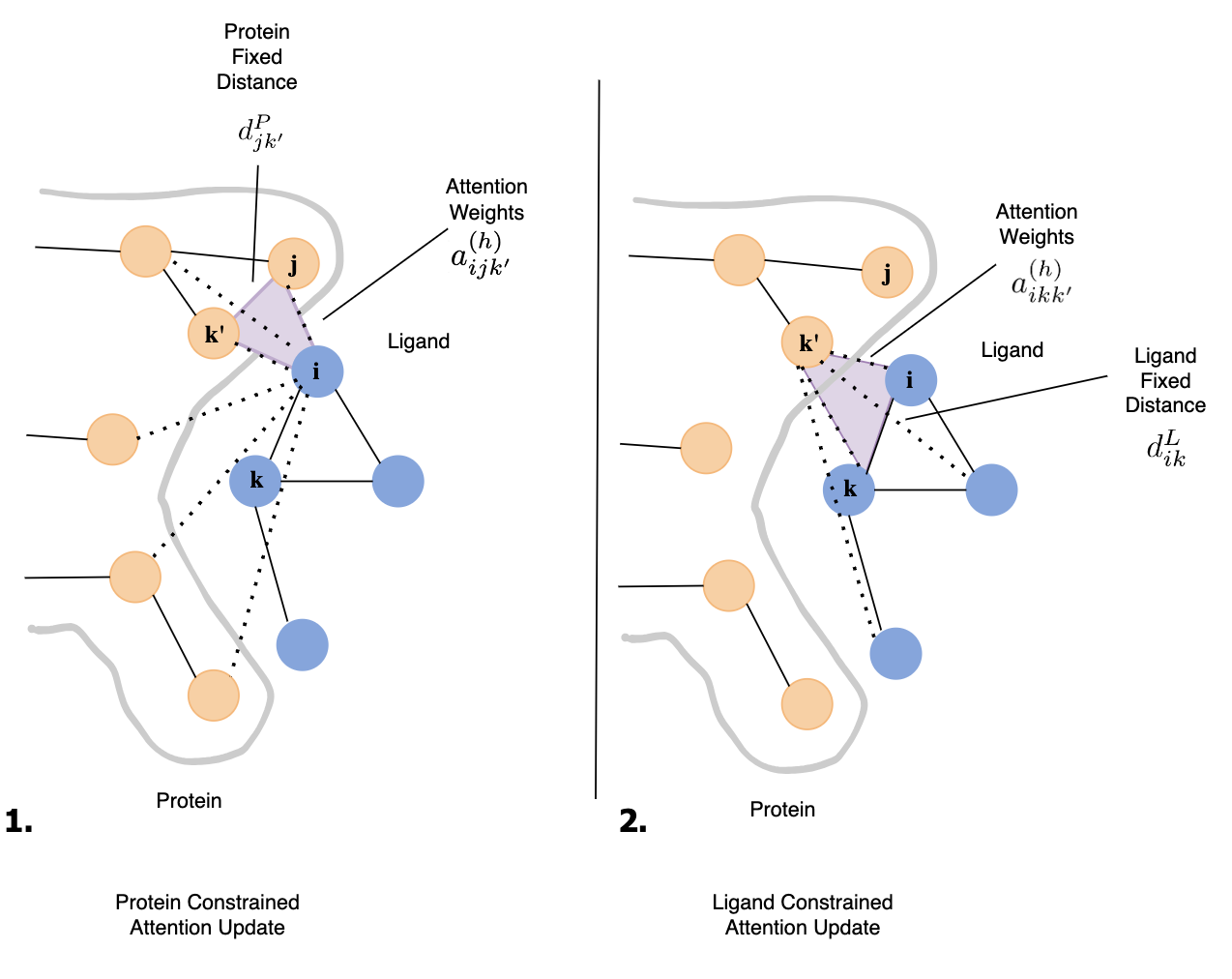}
         \caption{We learn nearby attention weights for the pairwise embeddings $h_{ij}^{PL}$ through a two-fold update process first (1.) based on intra-protein distances $d_{jk'}^{P}$ and secondly (2.) on intra-ligand distances $d_{ik}^{L}$. Performing these two updates ensures that we learn attention weights $a_{ijk'}^{(h)}$ for $h_{ij}^{PL}$ which respect both sets of distances, ensuring geometric-consistent information}
         \label{fig: results}
\end{figure}

The updated embeddings are processed through an MLP, $( h_{ij}^{new} = \text{MLP}(h_{ij}^{PL}) )$. We then perform standard multi headed attention on $h_{{v_i}^p}$ and $h_i^{L(N)}$ with $h_{ij}^{new}$ as a bias term, before updating pairwise embeddings with the updated $h_{{v_i}^p}$ and $h_i^{L(N)}$. The trioformer block then outputs the sets of ligand, protein and pairwise embeddings. After $N$ layers of we then take the ligand node embedding as it has sufficient pocket information. We generate our global graph embedding by taking the global avg pooling of the node embedding
\ref{eq: graph embedding formula}:
\begin{equation}
    \mathbf{g}^{L(N)} = AvgPool\left(\left\{\mathbf{h}^{L(N)}_i\right\}\right)
    \label{eq: graph embedding formula}
\end{equation}
We obtain the edge embedding by performing the same concatenation as TacoGFN.
\ref{eq: edge embedding formula}:
\begin{equation}
    \mathbf{e}_{ij}^{L(N)} = \mathbf{h}^{L(N)}_i + \mathbf{h}^{L(N)}_j
    \label{eq: edge embedding formula}
\end{equation}

\section{Experiments and Results} 
We evaluated our model using the CrossDock-100k dataset, a commonly used benchmark in the field, previously utilized by generative SBDD models \cite{Guan20233DEquivariant,Peng2022Pocket2Mol,Luo20223DGenerativeModel,Zhang2023LearningSubpocket,Zhang2023MoleculeGeneration,Shen2023TacoGFN}. 
Furthermore, we employ the docking score predictor developed by Shen et al. (2023) \cite{Shen2023TacoGFN}, which was trained on the ZINCDock-15M dataset. This choice ensures a fair comparison with their TacoGFN model, which relies on this predictor for affinity estimation. The ZINCDock-15M dataset comprises approximately 15 million docking simulation data points, generated by docking 1,000 random molecules from the ZINC20 database \cite{irwin2020zinc20} into each of the 15,207 unique pockets from the CrossDock-100k training split using QVina\cite{Alhossary2015Fast}. 

We evaluate the performance of our method using the same metrics employed by TacoGFN to assess drug candidates. In which we generate 100 unique molecules for each of the 100 protein pockets from the CrossDock-100k test set. The metrics we used are (1) \textbf{Diversity}, which is the average pairwise fingerprint Tanimoto distance between molecules across the 100 pockets; (2) \textbf{Docking Score}, an estimate of binding affinity between a generated molecule and its target pocket (3) \textbf{Top 10 Docking Score}, The average of the top 10 docking scores. The docking score is calculated using QVina2 \cite{Alhossary2015Fast}.

For the experiments, both the TacoGFN-Baseline and TacoGFN-Trioformer models underwent identical training, consisting of 10,000 steps on the CrossDock-100K training split. This training focused on optimizing the docking score predictor. Our evaluation aimed to determine whether incorporating geometric information led to improvements in terms of binding affinity for the target protein pocket. Each model was trained twice, and the best performing model was selected for each. Subsequently, we allowed each model to generate five sets of molecules.
\subsection{Results}

We conducted a series of experiments to assess the diversity, average docking score, and average top 10 docking score of the generated molecules. Our results show that we outperform the previous baseline in terms of docking score, suggesting that the introduction of additional geometrical information is beneficial for optimizing the generation of molecules with high binding affinity to the target protein pocket. However, there were minimal variations in diversity metrics, with the modified TacoGFN-Trioformer performing marginally worse than the baseline model. This outcome suggests that the geometric enhancements incorporated into the Trioformer might not directly influence the diversity of the molecular candidates in the way initially hypothesized.

Furthermore, we performed an ablation study focusing on the objective functions of docking, QED, and SA, to observe how the geometric information influences these key drug desirability metrics. We observed a similar increase in docking performance in the multi-objective setting, which further reinforces our hypothesis of the benefit of added protein-ligand geometric information.


\begin{table}[h]
    \vspace*{-6mm}
    \caption{ 
        Comparison of methods on Docking Score
    }
    \resizebox{\linewidth}{!}{
        \begin{tabular}{l|ccc}
            \toprule
             Objective   & Diversity ($\uparrow$) & DS ($\downarrow$)  & Top 10 DS ($\downarrow$)\\
            \midrule
            \textsc{TacoGFN-Baseline} & \textbf{0.58}$\pm$0.001 &  -9.00$\pm$0.01 & -11.41 $\pm$0.04 \\
             
            \textsc{TacoGFN-Trioformer} & 0.57$\pm$0.001 & \textbf{-9.34}$\pm$0.01 & \textbf{-11.85}$\pm$0.001 \\
            \bottomrule
        \end{tabular}
    }
    \label{tab: ablation}
    \vspace*{-6mm}
\end{table}

\subsection{Multi Objective Ablation}

We evaluated the influence of geometric embeddings in the multi-objective optimization scenario where the GFlowNet targeted the optimization of three objectives: (1) \textbf{Docking Score}, as previously described; (2) \textbf{QED}, a quantitative estimation of drug-likeness \cite{Bickerton2012}; and (3) \textbf{Synthetic Accessibility} (SA), an estimate of the ease of synthesis \cite{Ertl2009}. This experiment revealed that the performance of the TacoGFN-Trioformer model was largely similar to that of the TacoGFN-Baseline across the assessed metrics. TacoGFN-Trioformer continued to outperform the baseline in docking performance, indicating that the incorporation of trigonometrically consistent embeddings improves the model’s ability to predict binding affinities effectively. The SA scores between the two models exhibited negligible differences, implying that the additional spatial information integrated into the Trioformer does not adversely affect the synthetic feasibility of the generated molecules. However, there was a noticeable drop in QED, which may suggest that the spatial information led the model to prioritize this reward less.

\begin{table}[h!]
\centering
\begin{tabular}{c c c c c}
 \hline
  \multicolumn{1}{c}{Method} & \multicolumn{1}{c}{Metric} & \multicolumn{1}{c}{Avg.} & \multicolumn{1}{c}{Med} \\ [0.5ex]
 \hline\hline
 \multirow{4}{*}{TacoGFN-Baseline}
  & DS $\downarrow$& -7.75$\pm$0.006 & -7.73$\pm$0.005 \\
  & QED $\uparrow$& \textbf{0.69}$\pm$0.0007 & \textbf{0.69}$\pm$0.0008  \\
  & SA $\uparrow$& 0.79$\pm$0.0002 & -0.79$\pm$0.0003 \\
 \hline
 \multirow{4}{*}{TacoGFN-Trioformer}
  & DS $\downarrow$& \textbf{-7.94}$\pm$0.01 & \textbf{-7.94}$\pm$0.03 \\
  & QED $\uparrow$& 0.58$\pm$0.001 & 0.58$\pm$0.002 \\
  & SA $\uparrow$& \textbf{0.80}$\pm$0.0003 & \textbf{0.80}$\pm$0.0005 \\
 \hline
 \hline
\end{tabular}
\captionsetup{skip=10pt}
\caption{Ablation of GFlowNet Multiple Objectives}
\label{metrics}
\end{table}


\section{Conclusion} 
GFlowNets demonstrate a robust capacity to navigate and exploit molecular space, enhancing our ability to discover diverse drug candidates. In this study, we introduced trigonometrically consistent embeddings to GFlowNets, adding extra spatial pocket information. This modification was aimed at increasing the binding affinity of molecules generated for target pockets. Our results show that this additional information improves the binding affinity of the generated molecules to a pocket, underscoring the effectiveness of integrating geometric details into the pocket conditioning. However, our work has encountered certain limitations in regards to the diversity. Firstly, computational constraints limited our training to only 10,000 steps. Given the added complexity from the Trioformer model, more extensive training might have been beneficial. This limitation was primarily due to the memory constraints of our GPU setup. Another notable challenge was the potentially inadequate representation of intra-ligand distances. We currently employ a one-hot encoding strategy that only includes distances of direct neighbors. Future iterations could benefit from integrating shortest path distances, which would likely provide a more comprehensive spatial representation. Additionally, the specificity of the affinity predictor used may not have been adequately pocket-specific. A potential direction for future research could involve further normalizing the predicted docking scores, specifically adjusting for the average docking score across different pockets. This adjustment would refine the reward mechanism to penalize molecules that perform uniformly well across multiple pockets, encouraging more targeted molecule generation.

In conclusion, our work advances the field of structure-based drug design by integrating 3D geometric information about target protein pockets. By addressing these identified limitations, subsequent research could result in further improvements to enriching the pocket specificity of GFlowNets for SBDD.

\bibliographystyle{unsrt}
\bibliography{references}

\begin{thebibliography}{10}

\bibitem{Scannell2012}
Jack Scannell, Alex Blanckley, Helen Boldon, and Brian Warrington.
\newblock Diagnosing the decline in pharmaceutical r\&d efficiency.
\newblock {\em Nature Reviews Drug Discovery}, 11:191--200, 2012.

\bibitem{Bohacek1996StructureBased}
Regine~S. Bohacek, Colin McMartin, and Wayne~C. Guida.
\newblock The art and practice of structure-based drug design: A molecular modeling perspective.
\newblock {\em Medicinal Research Reviews}, 16(1):3--50, 1996.

\bibitem{Wang2005PDBbind}
Renxiao Wang, Xin Fang, Yong Lu, Chuan-yun Yang, and Shaomeng Wang.
\newblock The pdbbind database: methodologies and updates.
\newblock {\em Journal of Medicinal Chemistry}, 48(12):4111--4119, Jun 16 2005.

\bibitem{Guan20233DEquivariant}
Jiaqi Guan, Wesley~Wei Qian, Xingang Peng, Yufeng Su, Jian Peng, and Jianzhu Ma.
\newblock 3d equivariant diffusion for target-aware molecule generation and affinity prediction.
\newblock In {\em The Eleventh International Conference on Learning Representations}, 2023.

\bibitem{bengio2023gflownet}
Yoshua Bengio, Salem Lahlou, Tristan Deleu, Edward~J. Hu, Mo~Tiwari, and Emmanuel Bengio.
\newblock Gflownet foundations.
\newblock {\em arXiv preprint arXiv:2111.09266}, 2023.

\bibitem{Shen2023TacoGFN}
Tony Shen, Mohit Pandey, Jason Smith, Artem Cherkasov, and Martin Ester.
\newblock Tacogfn: Target conditioned gflownet for structure-based drug design.
\newblock {\em arXiv preprint arXiv:2310.03223}, 2023.

\bibitem{zhang2023e3bind}
Yangtian Zhang, Huiyu Cai, Chence Shi, Bozitao Zhong, and Jian Tang.
\newblock E3bind: An end-to-end equivariant network for protein-ligand docking.
\newblock {\em arXiv preprint arXiv:2210.06069}, 2023.

\bibitem{Peng2022Pocket2Mol}
Xingang Peng, Shitong Luo, Jiaqi Guan, Qi~Xie, Jian Peng, and Jianzhu Ma.
\newblock Pocket2mol: Efficient molecular sampling based on 3d protein pockets.
\newblock {\em arXiv preprint arXiv:2205.07249}, 2022.

\bibitem{Luo20223DGenerativeModel}
Shitong Luo, Jiaqi Guan, Jianzhu Ma, and Jian Peng.
\newblock A 3d generative model for structure-based drug design.
\newblock {\em arXiv preprint arXiv:2203.10446}, 2022.

\bibitem{Zhang2023MoleculeGeneration}
Zaixi Zhang, Yaosen Min, Shuxin Zheng, and Qi~Liu.
\newblock Molecule generation for target protein binding with structural motifs.
\newblock In {\em The Eleventh International Conference on Learning Representations}, 2023.

\bibitem{Zhang2023LearningSubpocket}
Zaixi Zhang and Qi~Liu.
\newblock Learning subpocket prototypes for generalizable structure-based drug design.
\newblock {\em arXiv preprint arXiv:2305.13997}, 2023.

\bibitem{pmlr-v202-guan23a}
Jiaqi Guan, Xiangxin Zhou, Yuwei Yang, Yu~Bao, Jian Peng, Jianzhu Ma, Qiang Liu, Liang Wang, and Quanquan Gu.
\newblock {D}ecomp{D}iff: Diffusion models with decomposed priors for structure-based drug design.
\newblock In {\em Proceedings of the 40th International Conference on Machine Learning}, volume 202 of {\em Proceedings of Machine Learning Research}, pages 11827--11846. PMLR, 23--29 Jul 2023.

\bibitem{Jain2023MultiObjectiveGFlowNets}
Moksh Jain, Sharath~Chandra Raparthy, Alex Hernandez-Garcia, Jarrid Rector-Brooks, Yoshua Bengio, Santiago Miret, and Emmanuel Bengio.
\newblock Multi-objective gflownets.
\newblock {\em arXiv preprint arXiv:2210.12765}, 2023.

\bibitem{Jumper2021}
John Jumper, Richard Evans, Alexander Pritzel, et~al.
\newblock Highly accurate protein structure prediction with alphafold.
\newblock {\em Nature}, 596:583--589, 2021.

\bibitem{Lu2022TANKBind}
Wei Lu, Qifeng Wu, Jixian Zhang, Jiahua Rao, Chengtao Li, and Shuangjia Zheng.
\newblock Tankbind: Trigonometry-aware neural networks for drug-protein binding structure prediction.
\newblock {\em bioRxiv}, 2022:495043, 2022.

\bibitem{jing2021learning}
Bowen Jing, Stephan Eismann, Patricia Suriana, Raphael J.~L. Townshend, and Ron Dror.
\newblock Learning from protein structure with geometric vector perceptrons.
\newblock {\em arXiv preprint arXiv:2009.01411}, 2021.

\bibitem{yun2020graph}
Seongjun Yun, Minbyul Jeong, Raehyun Kim, Jaewoo Kang, and Hyunwoo~J. Kim.
\newblock Graph transformer networks.
\newblock {\em arXiv preprint arXiv:1911.06455}, 2020.

\bibitem{riniker2015better}
Sereina Riniker and Gregory~A. Landrum.
\newblock Better informed distance geometry: Using what we know to improve conformation generation.
\newblock {\em Journal of Chemical Information and Modeling}, 55(12):2562--2574, 2015.

\bibitem{irwin2020zinc20}
John~J. Irwin, Khanh~G. Tang, Jennifer Young, Chinzorig Dandarchuluun, Benjamin~R. Wong, Munkhzul Khurelbaatar, Yurii~S. Moroz, John Mayfield, and Roger~A. Sayle.
\newblock Zinc20—a free ultralarge-scale chemical database for ligand discovery.
\newblock {\em Journal of Chemical Information and Modeling}, 60(12):6065--6073, 2020.

\bibitem{Alhossary2015Fast}
Amr Alhossary, Stephanus~Daniel Handoko, Yuguang Mu, and Chee-Keong Kwoh.
\newblock Fast, accurate, and reliable molecular docking with quickvina 2.
\newblock {\em Bioinformatics}, 31(13):2214--2216, July 2015.

\bibitem{Bickerton2012}
G.~Richard Bickerton, Gaia~V. Paolini, J{\'e}r{\'e}my Besnard, Sorel Muresan, and Andrew~L. Hopkins.
\newblock Quantifying the chemical beauty of drugs.
\newblock {\em Nature Chemistry}, 4(2):90--98, 2012.

\bibitem{Ertl2009}
Peter Ertl and Ansgar Schuffenhauer.
\newblock Estimation of synthetic accessibility score of drug-like molecules based on molecular complexity and fragment contributions.
\newblock {\em Journal of Cheminformatics}, 1(1):8, 2009.

\end{thebibliography}
\end{document}